\documentclass{article}

\usepackage{microtype}
\usepackage{graphicx}
\usepackage{subfigure}
\usepackage{booktabs} 

\usepackage[all,knot]{xy}

\xyoption{arc} 
\usepackage{amsfonts}
\usepackage{array}
\usepackage{euscript}
\usepackage{tikz}
\usepackage{epsfig, psfrag, epic}
\usepackage{graphicx}
\usepackage{xy}

\usepackage{pdfpages}

\usepackage{multirow}
\usepackage{amsmath}
\usepackage{enumerate}
\usepackage{color}
\usepackage{mathtools}
\usepackage{xcolor}
\usepackage[colorlinks=true,allbordercolors=white, citecolor=blue, breaklinks]{hyperref}

\usepackage{url}

\usepackage{breakurl}

\usepackage[utf8]{inputenc}

\input colordvi

\newcommand{\R}{{\mathbb R}}

\input colordvi                       

\usepackage{hyperref}

\usepackage[accepted]{arxivpaper}

\icmltitlerunning{Sorting Game}

\begin{document}

\twocolumn[
\icmltitle{Linear discriminant initialization for feed-forward neural networks}

\icmlsetsymbol{equal}{*}

\begin{icmlauthorlist}
\icmlauthor{Marissa Masden}{equal,uo}
\icmlauthor{Dev Sinha}{equal,uo}
\end{icmlauthorlist}

\icmlaffiliation{uo}{Department of Mathematics, University of Oregon, Oregon, USA}

\icmlcorrespondingauthor{Marissa Masden}{mmasden@uoregon.edu}
\icmlcorrespondingauthor{Dev Sinha}{dps@uoregon.edu}

\icmlkeywords{Machine Learning}

\vskip 0.3in
]



\printAffiliationsAndNotice{}  

\begin{abstract}
Informed by the basic geometry underlying feed forward neural networks, 
we initialize the weights of the first layer of a neural network using the linear discriminants which best 
distinguish individual classes. 
Networks initialized in this way take fewer training steps to reach the same level of training, and asymptotically have higher accuracy on training data. 
\end{abstract}


\section{Introduction}

We present an algorithm to find initial weights for 
networks that in a range of examples trains more effectively than randomly-initialized networks with the same architecture.  
Our results illustrate how geometry of a data set can inform the development of a 
network to be trained on that data.
We also expect that further development will prove useful to those working at the state of the art.

Effective methods for initializing the weights of deep networks \cite{He2015, orthogonal} allow for faster and more accurate training. 
Geometric and topological analyses of neural networks during training find that the first layer of a network eventually learns weights which match  ``features'' in the input space \cite{carlsson18}, and that extracting those features explicitly can be useful.

Here, we approximate these features of the data distribution via a process we call Linear Discriminant Sorting, or 
the ``Sorting Game,''  a deterministic method to initialize weights of a feedforward neural network. The weights which are found via the Sorting Game are then permitted to evolve during training, leading to greater flexibility. 

That initial, nonrandom weights affect a network's training has previous been shown in work on the lottery ticket hypothesis \cite{lotteryticket},  that large networks contain smaller subnetworks which train nearly as well as the original large network. Locating these smaller subnetworks requires the computationally-expensive process of weight pruning, and when one  reinitializes these subnetworks with random weights, they no longer perform well.    Here we find initial network weights which lead to a small neural network close to a near-optimal loss basin via a process which is less computationally intensive. Comparing with standard initializations shows improvement, especially when training networks with large batch sizes.

\begin{figure}[t]
	\centering
	\includegraphics[width=.5\columnwidth]{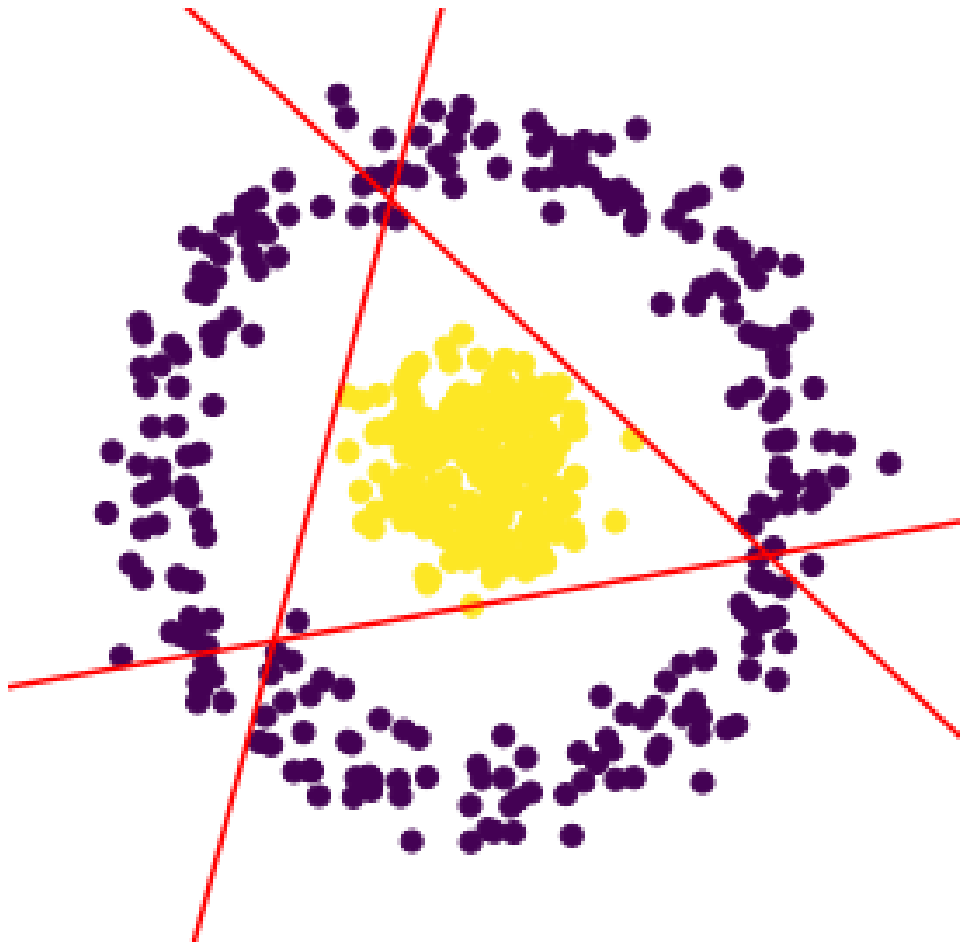}
	\caption{Hyperplanes representing the three neurons in the first layer of a small neural network trained on the annulus dataset, illustrating the relationship between the geometry of data and the first layer weights.}
	\label{Trained}
	\setlength{\belowcaptionskip}{-10pt}
\end{figure}

Through improvements in  performance, we provide evidence for a model for what some neural networks do, namely
 find discriminating features in early layers, and then use further layers to perform logic on those features.
The current algorithm and its implementation for relatively small networks, along with data sets which are generally modest 
-- though we do report in Section \ref{Alex} on the CIFAR-10 data set -- is also meant as a promising invitation to both scale up to larger networks and to 
implement for feedforward subnetworks of architectures such as transformers \cite{transformers}.  
More broadly, we see our main results as providing evidence
for the fruitfulness of  ideas from geometry and  topology to better understand and develop machine learning.



\section{Algorithm}

\subsection{Motivation and Background}

\begin{figure*}
	\begin{center}
		\centerline{\includegraphics[width=.7\textwidth]{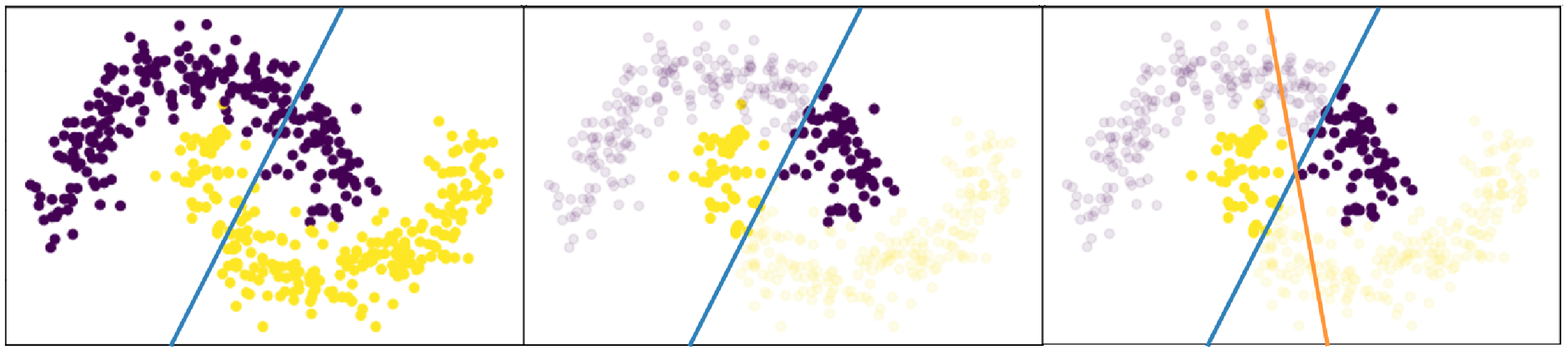}}
		\caption{The sorting game, in pictures. In (a), we compute the linear discriminant between the yellow and purple classes. In (b), we determine which points are unsorted by this discriminant, and in (c) we compute the linear discriminant between the remaining points. We recursively apply this process until all points are sorted.}
		\label{sorting-algorithm}
	\end{center}
\end{figure*}

The Linear Discriminant Sorting algorithm (informally, the ``sorting game'')
 builds on the mathematical description of neurons as hyperplanes partitioning the input space. In the sigmoid
setting, a neuron is effectively determined by a ``strip'' with a
  hyperplane at its center, on which the activation function changes values (typically from $-1$ to $1$).
In the ReLU setting, the activation function is constant on one side of the hyperplane and linear on the other. 

In some studies, the distribution of weights of the neurons in the first layer of trained network
 reflect the geometry of the dataset on which the network has been trained \cite{carlsson18}. In particular, we observe that in sigmoid networks trained on classification tasks, the hyperplanes representing the first layer neurons often appear to lie between the point clouds representing each class, as illustrated in Figure \ref{Trained}.

Our algorithm applies linear discriminant analysis \cite{fisher36lda, sklearn}  to compute hyperplanes best separating two classes of data. The unit vectors corresponding to those hyperplanes are then used to define first-layer neurons in a neural network. In our applications we primarily use this to initialize the first layer of weights, but we also initialize deeper fully-connected layers in a network following a fixed architecture  in Section \ref{Alex}. 


\subsection{Informal Description}


We describe the sorting game  applied to two classes of data, as 
additional classes are addressed by taking each class label $L$ and performing the sorting game  on 
``$L$ versus ${\sim}L$''. 



First, we find a hyperplane which separates the two classes by computing the linear discriminant between the data points in the input space. Then, we set the resulting components of the linear discriminant as a hyperplane for a neuron in the first layer of the network, with unit magnitude. 

We then discard the data points which have been sorted. To choose which points to discard, we first project the data onto the orthogonal complement of the hyperplane. We select a bias that maximizes the total number of data points which belong to opposing classes on opposite sides of the hyperplane,  which we then consider to be ``sorted.'' We remove the points which we consider sorted.  We then repeat the process of finding a linear discriminant, sorting and removing well-sorted points, until a unique linear discriminant cannot be computed. See Figure~\ref{sorting-algorithm}.   If there are multiple classes, we perform this procedure for the characteristic function of each class. 

We use these hyperplanes to initialize the first layer of a neural network, with at least as many neurons as hyperplanes found.  We then initialize any remaining layers of the network according to standard initialization schemes before training the network. We permit the discovered initial weights to evolve normally as part of the network.

\subsection{Formalized Algorithm}
\begin{algorithm}[H]
	\caption{Sorting Game Algorithm}
	\label{alg:example}
	\begin{algorithmic}
		\STATE {\bfseries Input:} Data points $X=\{\vec{x}_k\}$ where $\vec{x}_k \in {\mathbb R}^d$;\\
		 Labels $Y=\{y_k\}$;\\ Number of classes $n$.
		\STATE Initialize $j=0$.
		\FOR{$i=1$ {\bfseries to} $n$}
		\REPEAT
		\STATE Compute unit component vector $\vec{w}$ for the top linear discriminant on $x_k$ for the binary 
		class $\{ y_k = i\}$. 
		\STATE Store $\vec{w}$ as $w_j$.
		\STATE Set $z_k = \vec{w}\cdot x_k $
		\STATE Find bias $b \in \R$ maximizing the sum: $$\sum_k (y_k = i \text{ AND }z_k \leq b) \text{ OR }( y_k \neq i\text{ AND }z_k > b)$$
		\STATE Store $b$ as $b_j$. 
		\STATE Increment $j$.
		\STATE Remove points $(x_k, y_k)$ for all $k$ satisfying: \\$(y_k \neq i \text{ AND } z_k \leq b) \text{ OR } (y_k = i \text{ AND } z_k > b)$
		\UNTIL{$\# \{y_k: y_k = i\} < d \text{ OR } \# \{y_k: y_k \neq i\} < d $}
		\ENDFOR		
		\STATE Set network weights $W_{j,\textunderscore}^{(1)}= w_j$
	\end{algorithmic}
\end{algorithm}


\subsection{Sampling and Dimensional Reduction}

Performing the linear discriminant analysis on samples from the data  reduces computational expense.  We use
such a strategy in Section \ref{Alex}.  Less obvious but also quite helpful is decreasing dimensionality. 
If there are $N$ data point in $d$-dimensional space with $N>d$, it is $\mathcal{O}(Nd^2)$ 
to compute all linear discriminants \cite{ldacomplexity}. Instead, we may perform the linear discriminant analysis on a subset of input variables at a time. 
Doing so on $ \frac{d}{n}$ features of the input data set at a time (for $n$ times),  
 leads to $\mathcal{O}(Nd^2/n)$ complexity. 
This leads to a large practical speed up, for example, when fine-tuning the feedforward subnetwork of 
AlexNet on CIFAR-10 ub Section~\ref{Alex}. 


 % ...

\section{Results}

\begin{figure*}
	\centering
	\begin{tabular}{l|l|l|l|}
		\cline{2-4}
		
		& \multicolumn{1}{c|}{Batch Size 25}                                 & \multicolumn{1}{c|}{Batch Size 100}&\multicolumn{1}{c|}{Batch Size 500}\\ \hline
		\multicolumn{1}{|l|}{ $\eta = 0.001$} &  
		$12.0 \pm 1.2 ^*$ &
		$ 17.4 \pm 1.6^*$ &
		$29.0 \pm 1.8 ^*$	
		\\ 
		
		\multicolumn{1}{|l|}{ $\eta = 0.005$} & 
		$4.5 \pm1.2^*$ & 
		$7.6 \pm 1.7^*$ & 
		$15.2 \pm 1.9^*$
		
		\\ 
		\multicolumn{1}{|l|}{ $\eta = 0.01$}  & 
		$6.2\pm1.5^*$ & 
		$6.7 \pm 1.0^*$ & 
		$11.2 \pm 1.4^*$
		\\ \hline
	\end{tabular}
	
	\caption{Comparison of number of training epochs to threshold accuracy across batch size and learning rate $\eta$ for FashionMNIST data set. Displayed as $(\mu_{rand}-\mu_{lda})\pm SE$}
	
	\label{bigtable}
\end{figure*}

\begin{figure*}
	\centering
	%

	\begin{tabular}{l|l|l|l|}
		\cline{2-4}
		& \multicolumn{1}{c|}{Batch Size 25}                                 & \multicolumn{1}{c|}{Batch Size 100}&\multicolumn{1}{c|}{Batch Size 500}\\ \hline
		\multicolumn{1}{|l|}{ $\eta = 0.001$} 
		& $ 0.27\% \pm 0.046\% ^*$ 
		& $0.279\% \pm 0.049 \%^*$
		& $0.776\% \pm 0.078 \%^*$
		\\ 		
		\multicolumn{1}{|l|}{ $\eta = 0.005$} 
		& $0.032\% \pm 0.058\%$ 
		& $0.165\% \pm 0.055\%^*$ 
		& $0.392\% \pm 0.065\%^*$
		\\ 
		\multicolumn{1}{|l|}{ $\eta = 0.01$}  
		& $0.164\% \pm 0.061\%^*$
		& $0.169\% \pm 0.066\%^*$
		& $0.258\% \pm 0.051\%^*$
		\\ \hline  
	\end{tabular}
	
	\caption{Comparison of minimum validation error (in percent) across batch size and learning rate, following 100 training epochs. Displayed as $(Err_{rand}-Err_{lda})\pm SE$}
	\label{bigtable2}
\end{figure*}

We compare networks initialized with the LDA sorting game to those initialized randomly. In most experiments, we use the LDA Sorting algorithm to 
determine a number of neurons to initalize deterministically, and then create both LDA-initialized and entirely randomly-initialized networks with the same architecture.  Results with a priori fixed architecture are in Section \ref{Alex}.

We compare the training performance between the two initialization schemes both visually, comparing epoch vs. accuracy graphs over many trials,
and using three metrics. The first metric is the difference between $\mu_{lda}$ and $\mu_{rand}$, where $\mu$ is the average number of training steps needed to for training error to reach \textit{threshold accuracy}. Here, the \textit{threshold accuracy} is defined as the maximum observed training accuracy of the least-accurate network trained with the same hyperparameters. We define the threshold accuracy in this way to allow for consistent meaning across hyperparameters.  
The second metric is the difference in minimum validation error between the LDA-initialized networks $Err_{lda}$ and the randomly-initialized networks $Err_{rand}$. Lastly, when practical, we determine how much larger a randomly-initialized network must be to reach the same performance as a small network initialized only with LDA-initialized neurons.

These measurements capture the improvements in performance of the sorting game algorithm.  
When we see below that 
$\mu_{lda}$ is significantly less than $\mu_{rand}$, then LDA-initialized networks reach a given training accuracy sooner than those initialized randomly.
When we see that the minimum validation error $Err_{lda}$ is less than that of $Err_{rand}$ this indicates that LDA sorting  leads to better generalization by the trained network. Lastly, if a much larger network is necessary in order for a randomly-initialized network to perform similarly to a LDA-initialized network, this indicates that LDA sorting does find small networks which perform as well as larger networks. 

\subsection{Sigmoid Activation}
Our first case is that of 
sigmoid-activated networks. We  compare networks with LDA-sorted first layers against networks with the same architecture and orthogonal weight initialization \cite{orthogonal}. 

\begin{figure}[H]
	\includegraphics[width=\columnwidth]{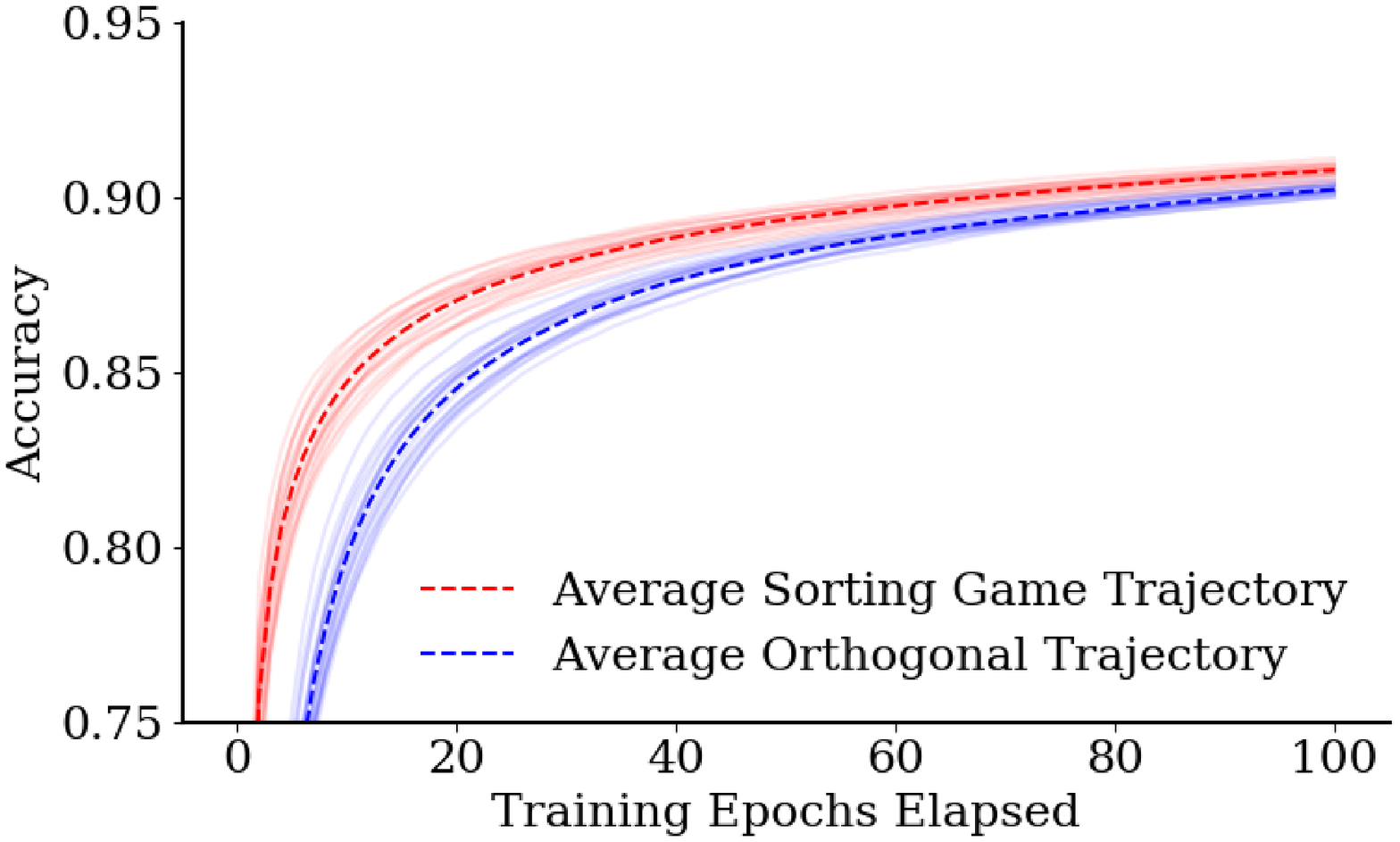}
	\includegraphics[width=\columnwidth]{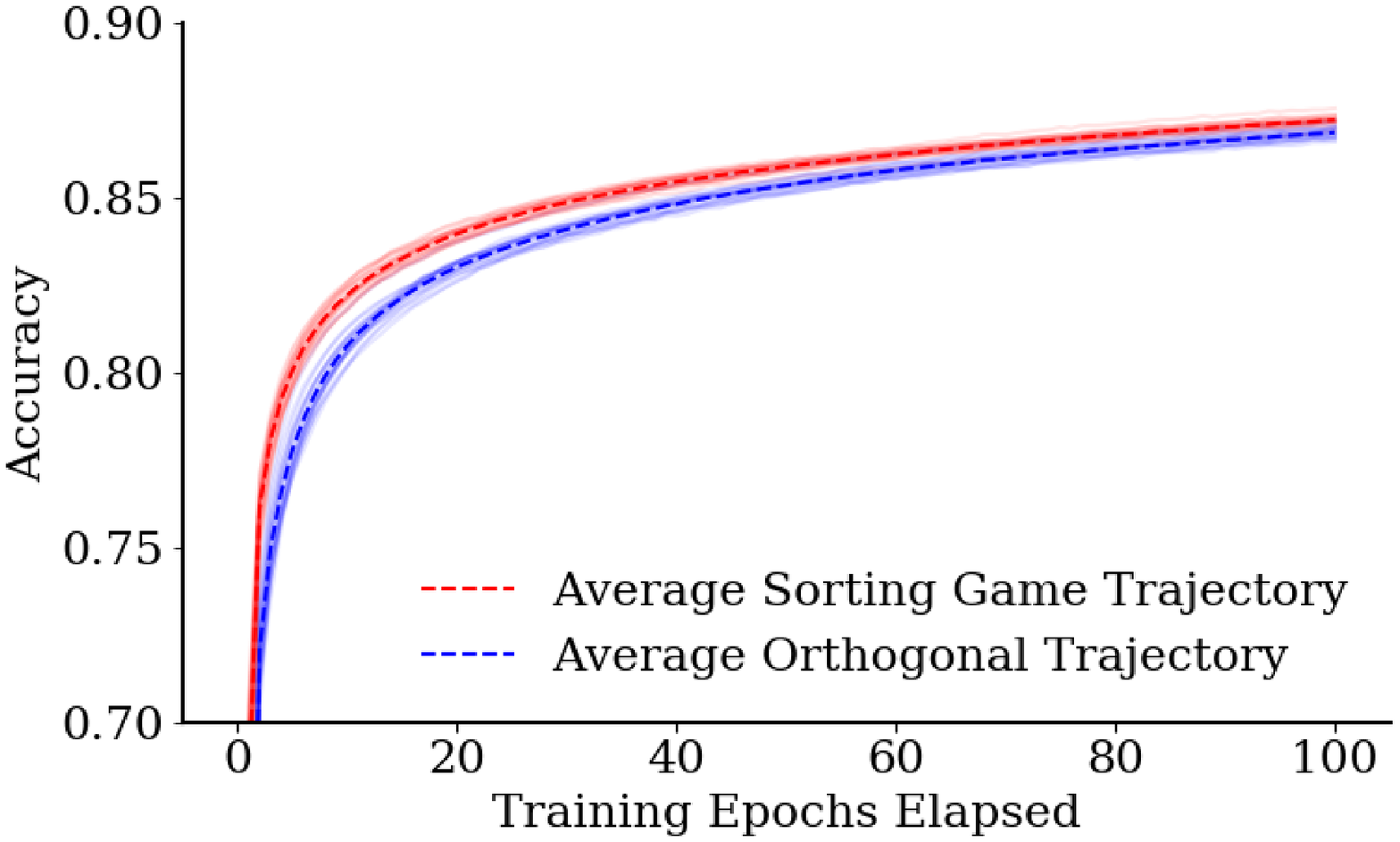}
	\caption{ Training accuracy plotted through the training of 20 different fully-connected feedforward neural networks on the MNIST dataset (top) and the Fashion MNIST dataset (bottom).} 
	\label{MNIST}
\end{figure}

For the MNIST dataset, the LDA initialization algorithm finds 21 weights, which we use to initialize 21 hidden units. Comparing the training trajectory of networks (784 input neurons, 21 hidden neurons, 10 output neurons, softmax and categorical crossentropy loss) initialized with these 21 components against randomly-initialized networks of the same architecture, the LDA-initialized networks reach higher training accuracy significantly sooner than those initialized entirely randomly in all but the networks trained with a very low batch size and high learning rate.  Visually, in Figure~\ref{MNIST} we see that the accuracy of the
LDA-initialized networks (in red, in all figures) are consistently higher than the randomly initialized networks (in blue, in all figures).  
We observe similar results for the Fashion MNIST dataset, where the LDA initialization algorithm finds 28 components. We initialized a network with 28 hidden units, 10 output units, and a softmax output layer, and trained it using stochastic gradient descent and categorical crossentropy loss.

 \begin{figure}
 	\includegraphics[width=\columnwidth]{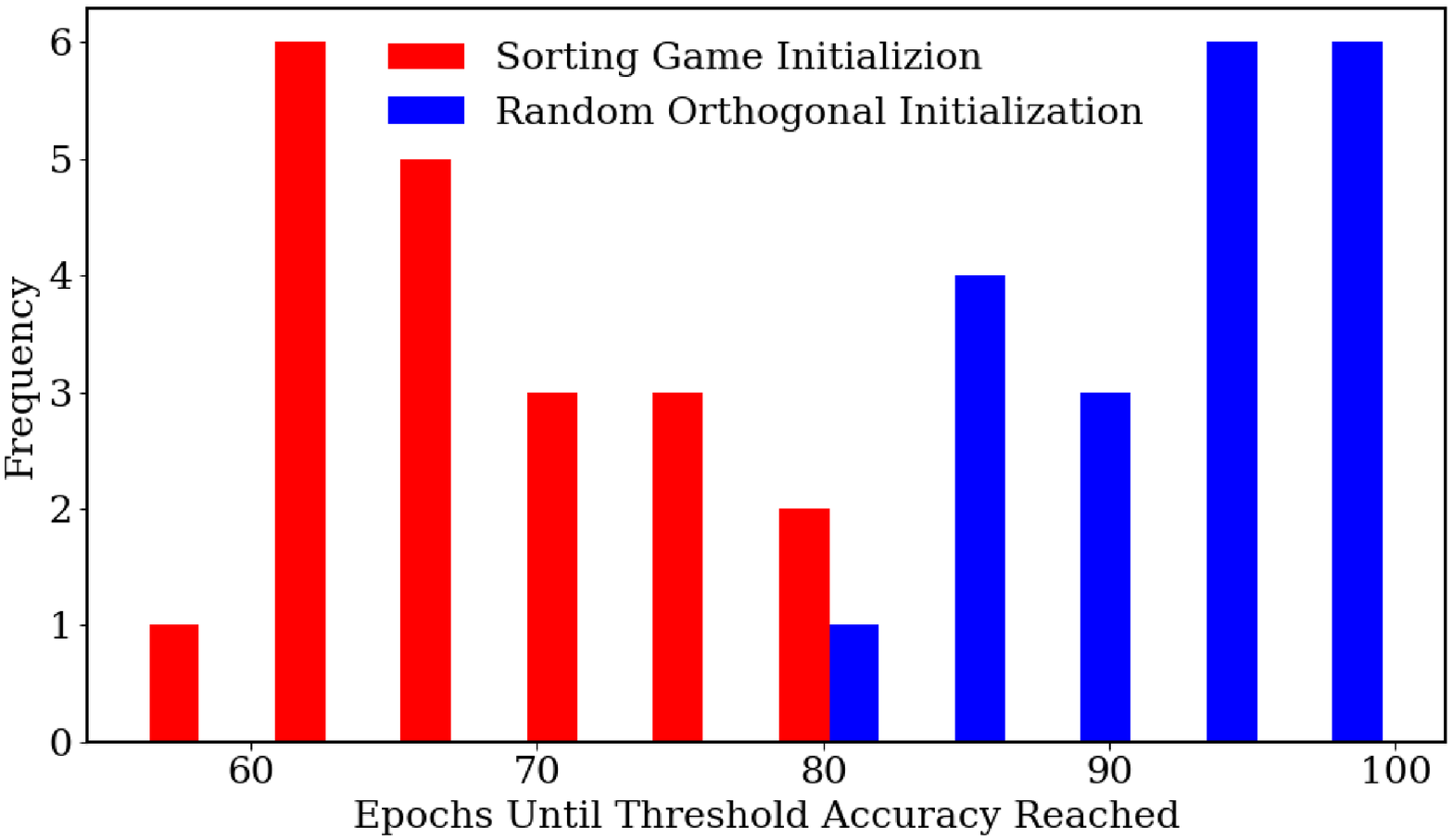}
 	
 	\caption{Distribution of the number of epochs until threshold accuracy is reached, over 20 training sequences of 100 epochs each, on MNIST Dataset.}
 \end{figure}
 
 \begin{figure}
 	\includegraphics[width=\columnwidth]{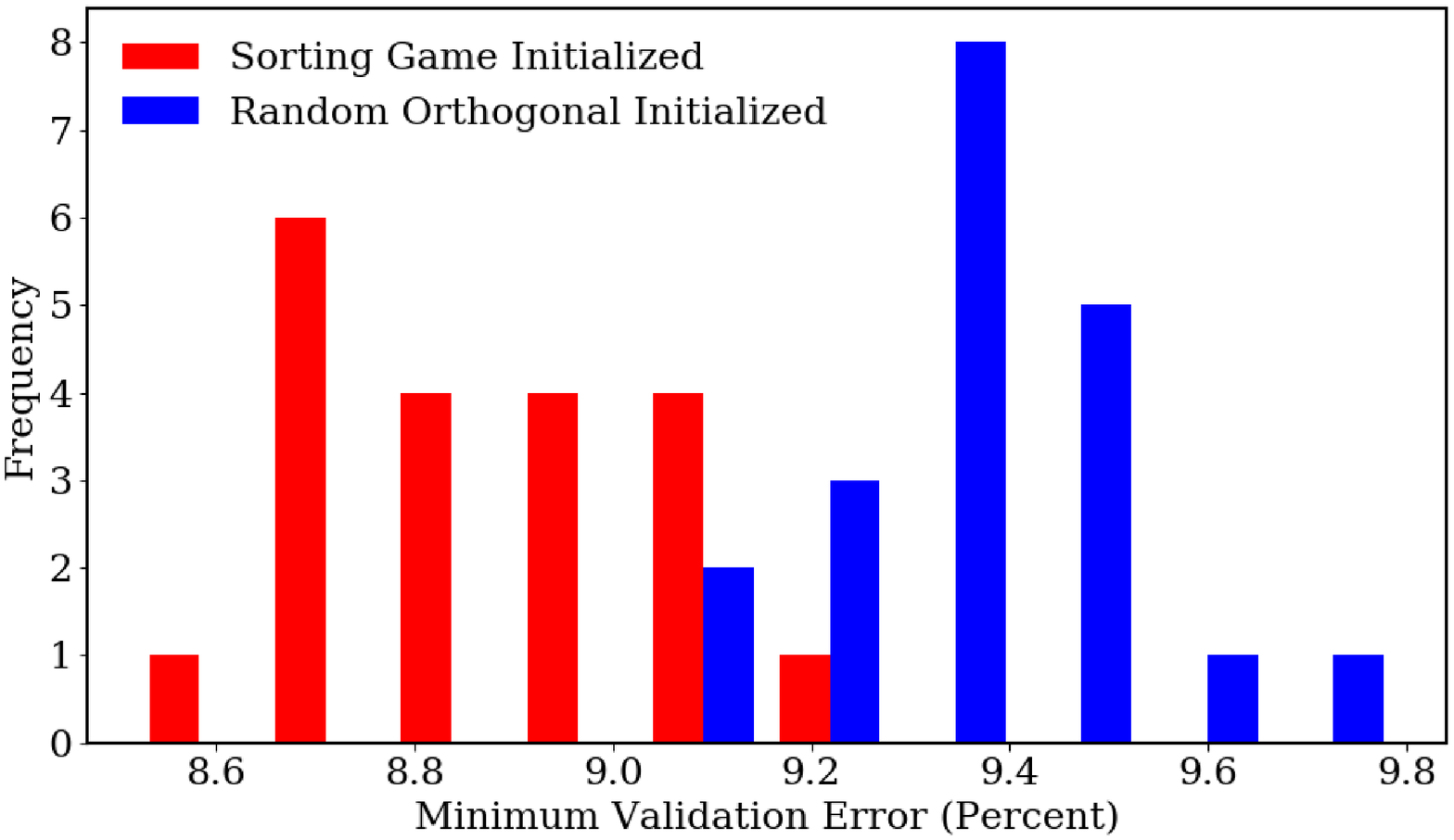}
 	
 	\caption{Example of the distribution of minimum validation error after 100 epochs of training. (Training on MNIST dataset)}
 \end{figure}

\subsection{Comparison Across Batch Size and Learning Rate}
To ensure that the improved training we see from LDA Initialization is robust, we performed the same experiment across batch sizes and learning rates for the MNIST and FashionMNIST dataset initializations. We keep the initialized (sorted) weights the same but independently randomize the remaining weights. The tables in Figures~\ref{bigtable} and \ref{bigtable2} demonstrate the comparison between the behavior of LDA-sorted networks and those with random initialization. 
We consistently see substantial differences in the number of epochs required to reach threshold accuracy, namely about ten epochs out of ninety, and small but significant differences in the minimum validation error. In both tables, effect size generally increases with increasing batch size, and decreases with increased learning rate.

Lastly, we consider what size of network is necessary for a completely randomly-initialized network to match the performance of the 28-hidden-unit network whose first layer is completely LDA-initialized on FashionMNIST. This again varies by batch size and learning rate. With a low batch size (25) and high learning rate ($\eta=.01$), the LDA-initialized networks perform as well as a randomly-initialized network of about 1.5 times the size (42 neurons). However, with a large batch size (500) and low learning rate ($\eta=.001$), the LDA-initialized networks perform as well as a randomly-initialized network of roughly 4 times the size (112 neurons). The effect size again follows a similar pattern as before.

\subsection{Initializing a Subset of Neurons} 
In the case where architecture is pre-selected, the sorting game still gives a benefit to training behavior. Using LDA sorting to initialize only a subset of the first layer's weights, and then randomly initializing the remaining weights, continues to demonstrate improved training performance over orthogonal initialization, though the improvement diminishes as additional neurons are added, as in Figure \ref{extraneurons}.

\begin{figure}
	\begin{tabular}{l|l|l|}
		\cline{2-3}                              & \multicolumn{1}{c|}{$\mu_{rand}-\mu_{lda}$}&\multicolumn{1}{c|}{ $Err_{rand}-Err_{lda}$}\\ \hline
		\multicolumn{1}{|l|}{$\times0$ Extra Neurons} &  
		$29.1\pm 1.7$ & 
		$0.81\% \pm 0.14\%$ 
		\\ 
		
		\multicolumn{1}{|l|}{$\times1$ Extra Neurons } & 
		$19.0\pm 1.3$ & 
		$0.35\% \pm 0.06\%$ 		
		\\ 
		
		\multicolumn{1}{|l|}{ $\times2$ Extra Neurons}  & 
		$12.7 \pm 1.3$ & 
		$0.24\%\pm 0.05\%$
		\\ 
			\multicolumn{1}{|l|}{ $\times3$ Extra Neurons}  & 
		$11.7 \pm 1.5$ & 
		$0.28 \% \pm 0.06\%$ 
		\\ 
		\multicolumn{1}{|l|}{ $\times4$ Extra Neurons}  & 
		$11.0 \pm 1.3$ &
		$0.25\% \pm 0.06\%$
		\\\hline
	\end{tabular}
	\caption{ Performance of progressively larger networks trained on Fashion MNIST for 100 epochs. The Sorting Game initialization finds 28 neurons, and training networks with exactly 28 hidden neurons is represented in the first row. Each subsequent row represents a larger network with 28$\times n$ ``extra neurons" which are randomly initialized, in addition to the Sorting Game initialized subnetwork.}
	\label{extraneurons}
\end{figure}


\subsection{AlexNet Fine Tune} \label{Alex}
\begin{figure}
	\includegraphics[width=\columnwidth]{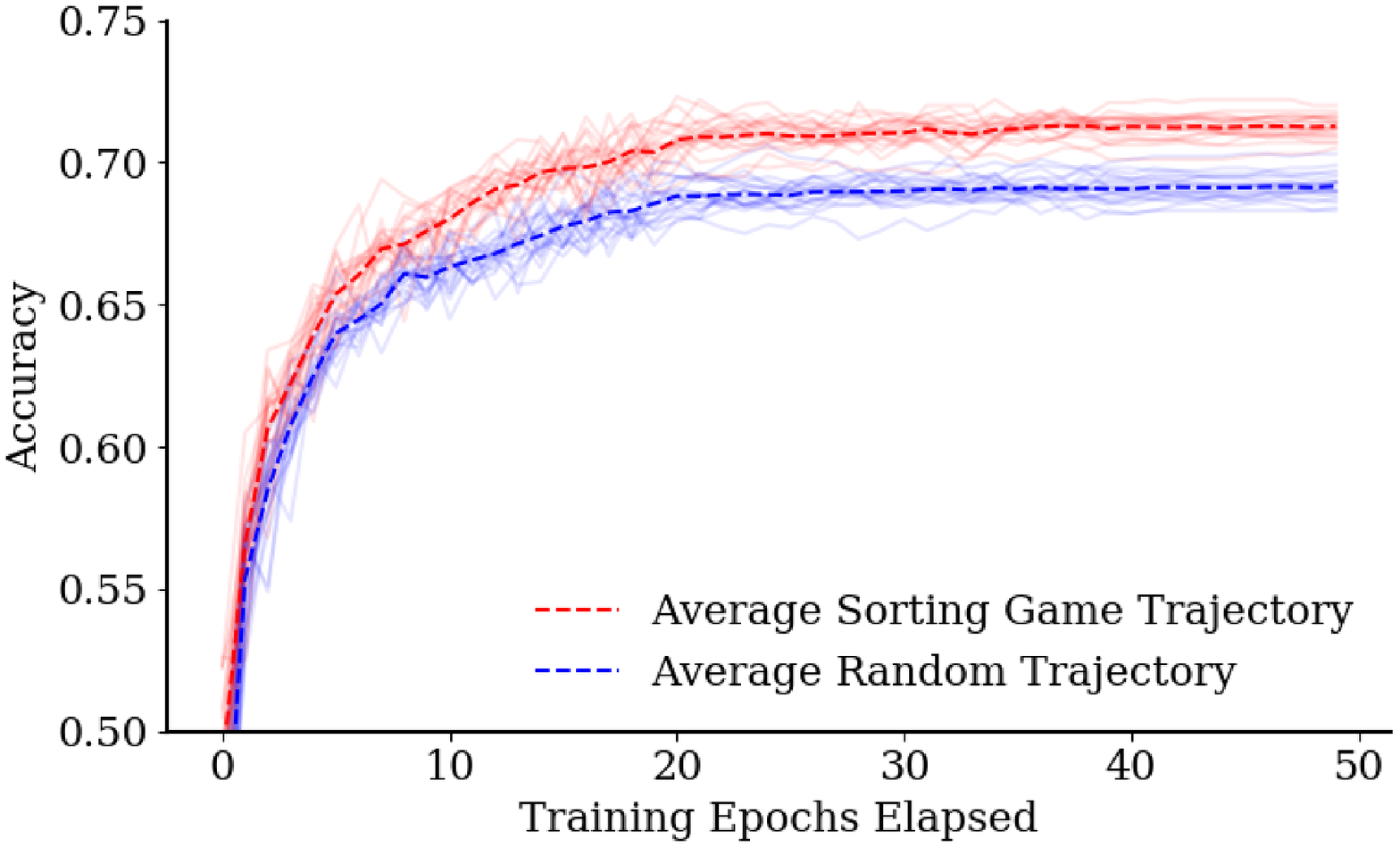}
	\caption{Validation accuracy throughout training when fine tuning an AlexNet implementation with fixed convolutional layers and mutable feedforward layers with sigmoid activation. Comparison of training between LDA-initialized first feedforward layer, and Xavier Gaussian-initialized.}
	\label{alexnet}
\end{figure}

We use the sampling modification described in Sec. 2.3 to initialize 1048 neurons using the output of AlexNet convolutional layers \cite{alexnet}. 
We use the CIFAR-10 dataset \cite{cifar10}, and resize images to the appropriate size for input into the AlexNet convolutional layers. 
We  train a feedforward network with 4092 input neurons, 1024 hidden neurons in the first layer, and 10 output neurons with softmax. 
We then followed a learning rate schedule with initial learning rate of .01 and a learning rate decay factor of 0.7 every 10 epochs, with a dropout factor of 0.4.  Compared to a Gaussian-initialized network, the linear discriminant initialization leads to significant improvement in initial training, as seen in Figure \ref{alexnet}. Since training was performed on data augmented by random affine transformations, training accuracy was inconsistent. Instead, we compute threshold accuracy and training time on validation data.  During a 50-epoch training run, Sorting Game initialized networks reached threshold accuracy on average $7.2$ epochs sooner than Xavier Normal-initialized networks \cite{xavier}. Additionally, Sorting Game-initialized networks reached an average of $2.13$ percentage points lower minimum validation error ($95\%$ CI $1.86$ to $2.41$ percentage points). 


While the sorting game was designed to handle sigmoid activation functions, an identical experiment with the same weight initialization was also performed with the remaining feedforward layers of AlexNet with ReLU activation.  Compared to He initialization \cite{He2015}, the training still appeared improved, but the difference was less pronounced. 

\subsection{Global Performance, and Deeper Layers}

In practice, the amount of computational time it takes to run this algorithm is lower than that of pruning a large network, 
but higher than that of running a randomly-initialized network a bit longer. On one machine, applying the (non-optimized) Sorting Game algorithm on the MNIST dataset takes approximately 170 seconds, but a single epoch of training what is now considered a fairly small network with 800 hidden units such as those used in \cite{oldmnist} on the same device takes about 17 seconds. Training a full sized network to completion, roughly one hundred epochs, in order to prune its weights would thus take 
approximately ten times  as long as Sorting Game initialization.

Finally, we  report that naively applying linear discriminant analysis
 to the image of data under the first layer, in order to initialize a 
second layer, did not yield positive results.  At the moment, the Sorting Game only has strong supporting evidence
as a way to initialize the first layer.



 % ...

\section{Discussion}
\subsection{Interpretation}
Our experiments demonstate improvement in training performance when using LDA initialization compared to 
standardly utilized randomized initializations. 
This improvement is robust across hyperparameters and also occurs in larger architectures. 

We expect that, with further optimization, this algorithm could be of value to machine learning practitioners.
But this work may be of greater theoretical significance in that it sheds light on 
the geometry of the loss landscapes of neural network training. Because lower stochasticity (large batch size and lower learning rate) leads to a greater separation between Sorting Game-initialized networks and those networks which are randomly initialized, a reasonable interpretation of these results is that LDA Sorting finds a loss basin which has an optimum closer to a global optimum 
than a randomly-initialized network.  We thus have a deterministic algorithmic step which could be incorporated in a number
of ways to achieve a combination of higher accuracy and less computational expense. 

In some applications larger batch sizes are desirable for more efficient parallel computation when training \cite{largebatch}. 
However, large batch training has pitfalls such as, potentially, decreased generalization \cite{badlargebatch}. 
Since the improvements the Sorting Game are more pronounced when training with larger batch sizes, 
 this initialization scheme should lead to large batch sizes being more feasible in practice.

\subsection{Future Directions}

Our primary goal is to find geometric understanding of small neural networks which train comparatively well, which we have
identified in the first layer.  We conjecture that further layers could be addressed not through sorting but through a modification
of an algorithm such as Adaboost  \cite{adaboost}, which generates a final classification result via linear combination.  
Unlike  Adaboost, we permit  initial weights to evolve via a (potentially deep) neural network's training process.
But it may be possible to combine efforts, so to speak, and apply multiclass Adaboost \cite{multi-adaboost} to the outputs of the 
linear discriminants which are found via the Sorting Game, to initialize all or part of deeper layers. 

Additionally, though our results were more strongly supported for sigmoid activation functions than ReLu, we 
believe that the general principles of its initialization scheme are applicable in a broader scope.
Some modification will be needed to be  applicable for varied classes of activation functions, which opens up
an avenue for inquiry, namely the interplay between activation functions, geometry of data, and geometry of trained networks.
We also expect that other network architectures can be initialized via similar methods, and in particular expect
 the Sorting Game to be applicable to fully-connected feedforward portions of recurrent architectures, such 
 as transformers. 

Ultimately, modifying the Sorting Game so that it can be fruitfully applied to multiple layers in a network 
 would not only be of greater practical value,  but is likely to require deeper insight into the geometry of data, networks and loss landscapes. 


\bigskip


 % ...

\bibliography{bibliography}

\begin{thebibliography}{16}
\providecommand{\natexlab}[1]{#1}
\providecommand{\url}[1]{\texttt{#1}}
\expandafter\ifx\csname urlstyle\endcsname\relax
  \providecommand{\doi}[1]{doi: #1}\else
  \providecommand{\doi}{doi: \begingroup \urlstyle{rm}\Url}\fi

\bibitem[Cai et~al.(2008)Cai, He, and Han]{ldacomplexity}
Cai, D., He, X., and Han, J.
\newblock {Training linear discriminant analysis in linear time}.
\newblock In \emph{Proceedings - International Conference on Data Engineering},
  pp.\  209--217, 2008.
\newblock ISBN 9781424418374.
\newblock \doi{10.1109/ICDE.2008.4497429}.

\bibitem[Carlsson \& Gabrielsson(2018)Carlsson and Gabrielsson]{carlsson18}
Carlsson, G. and Gabrielsson, R.~B.
\newblock {Topological Approaches to Deep Learning}.
\newblock In \emph{Topological Data Analysis}, pp.\  119--146. Springer, 2018.
\newblock URL \url{http://arxiv.org/abs/1811.01122}.

\bibitem[Fisher(1936)]{fisher36lda}
Fisher, R.~A.
\newblock The use of multiple measurements in taxonomic problems.
\newblock \emph{Annals of Eugenics}, 7\penalty0 (7):\penalty0 179--188, 1936.

\bibitem[Frankle \& Carbin(2019)Frankle and Carbin]{lotteryticket}
Frankle, J. and Carbin, M.
\newblock {The lottery ticket hypothesis: Finding sparse, trainable neural
  networks}.
\newblock In \emph{7th International Conference on Learning Representations,
  ICLR 2019}, pp.\  1--42, 2019.

\bibitem[Freund \& Schapire(1997)Freund and Schapire]{adaboost}
Freund, Y. and Schapire, R.~E.
\newblock A decision-theoretic generalization of on-line learning and an
  application to boosting.
\newblock \emph{Journal of computer and system sciences}, 55\penalty0
  (1):\penalty0 119--139, 1997.

\bibitem[Glorot \& Bengio(2010)Glorot and Bengio]{xavier}
Glorot, X. and Bengio, Y.
\newblock {Understanding the difficulty of training deep feedforward neural
  networks}.
\newblock 9:\penalty0 249--256, 2010.
\newblock ISSN 15324435.

\bibitem[Hastie et~al.(2009)Hastie, Rosset, Zhu, and Zou]{multi-adaboost}
Hastie, T., Rosset, S., Zhu, J., and Zou, H.
\newblock Multi-class adaboost.
\newblock \emph{Statistics and its Interface}, 2\penalty0 (3):\penalty0
  349--360, 2009.

\bibitem[He et~al.(2015)He, Zhang, Ren, and Sun]{He2015}
He, K., Zhang, X., Ren, S., and Sun, J.
\newblock {Delving deep into rectifiers: Surpassing human-level performance on
  imagenet classification}.
\newblock In \emph{Proceedings of the IEEE International Conference on Computer
  Vision}, volume 2015 Inter, pp.\  1026--1034, 2015.
\newblock ISBN 9781467383912.
\newblock \doi{10.1109/ICCV.2015.123}.

\bibitem[Keskar et~al.(2019)Keskar, Nocedal, Tang, Mudigere, and
  Smelyanskiy]{badlargebatch}
Keskar, N.~S., Nocedal, J., Tang, P. T.~P., Mudigere, D., and Smelyanskiy, M.
\newblock {On large-batch training for deep learning: Generalization gap and
  sharp minima}.
\newblock In \emph{5th International Conference on Learning Representations,
  ICLR 2017 - Conference Track Proceedings}, pp.\  1--16, 2019.

\bibitem[Krizhevsky \& Hinton(2009)Krizhevsky and Hinton]{cifar10}
Krizhevsky, A. and Hinton, G.
\newblock Learning multiple layers of features from tiny images.
\newblock 2009.

\bibitem[Krizhevsky et~al.(2012)Krizhevsky, Sutskever, and Hinton]{alexnet}
Krizhevsky, A., Sutskever, I., and Hinton, G.~E.
\newblock Imagenet classification with deep convolutional neural networks.
\newblock In Pereira, F., Burges, C. J.~C., Bottou, L., and Weinberger, K.~Q.
  (eds.), \emph{Advances in Neural Information Processing Systems 25}, pp.\
  1097--1105. Curran Associates, Inc., 2012.
\newblock URL \url{https://tinyurl.com/y2u8t3b6}.

\bibitem[Lucas et~al.(2003)Lucas, Panaretos, Sosa, Tang, Wong, and
  Young]{oldmnist}
Lucas, S.~M., Panaretos, A., Sosa, L., Tang, A., Wong, S., and Young, R.
\newblock Icdar 2003 robust reading competitions.
\newblock In \emph{Proceedings of the Seventh International Conference on
  Document Analysis and Recognition - Volume 2}, ICDAR ’03, pp.\  682, USA,
  2003. IEEE Computer Society.
\newblock ISBN 0769519601.

\bibitem[Pedregosa et~al.(2011)Pedregosa, Varoquaux, Gramfort, Michel, Thirion,
  Grisel, Blondel, Prettenhofer, Weiss, Dubourg, Vanderplas, Passos,
  Cournapeau, Brucher, Perrot, and {{\'E}}douard Duchesnay]{sklearn}
Pedregosa, F., Varoquaux, G., Gramfort, A., Michel, V., Thirion, B., Grisel,
  O., Blondel, M., Prettenhofer, P., Weiss, R., Dubourg, V., Vanderplas, J.,
  Passos, A., Cournapeau, D., Brucher, M., Perrot, M., and {{\'E}}douard
  Duchesnay.
\newblock Scikit-learn: Machine learning in python.
\newblock \emph{Journal of Machine Learning Research}, 12\penalty0
  (85):\penalty0 2825--2830, 2011.
\newblock URL \url{http://jmlr.org/papers/v12/pedregosa11a.html}.

\bibitem[Saxe et~al.(2014)Saxe, McClelland, and Ganguli]{orthogonal}
Saxe, A.~M., McClelland, J.~L., and Ganguli, S.
\newblock {Exact solutions to the nonlinear dynamics of learning in deep linear
  neural networks}.
\newblock In \emph{2nd International Conference on Learning Representations,
  ICLR 2014 - Conference Track Proceedings}, pp.\  1--22, 2014.

\bibitem[Smith et~al.(2018)Smith, Kindermans, Ying, and Le]{largebatch}
Smith, S.~L., Kindermans, P.~J., Ying, C., and Le, Q.~V.
\newblock {Don't decay the learning rate, increase the batch size}.
\newblock \emph{6th International Conference on Learning Representations, ICLR
  2018 - Conference Track Proceedings}, \penalty0 (2017):\penalty0 1--11, 2018.

\bibitem[Vaswani et~al.(2017)Vaswani, Shazeer, Parmar, Uszkoreit, Jones, Gomez,
  Kaiser, and Polosukhin]{transformers}
Vaswani, A., Shazeer, N., Parmar, N., Uszkoreit, J., Jones, L., Gomez, A.~N.,
  Kaiser, {\L}., and Polosukhin, I.
\newblock {Attention is all you need}.
\newblock In \emph{Advances in Neural Information Processing Systems}, volume
  2017-Decem, pp.\  5999--6009, 2017.

\end{thebibliography}
\bibliographystyle{icml2020}

\end{document}